# Virtual information system on working area

Spits Warnars H.L.H
Department of Computing and Mathematics
Manchester Metropolitan University
S.Warnars@mmu.ac.uk

*Abstract* : In order to get strategic positioning for competition in business organization, the information system must be ahead in this information age where the information as one of the weapons to win the competition and in the right hand the information will become a right bullet. The information system with the information technology support isn't enough if just only on internet or implemented with internet technology. The growth of information technology as tools for helping and making people easy to use must be accompanied by wanting to make fun and happy when they make contact with the information technology itself. Basically human like to play, since childhood human have been playing, free and happy and when human grow up they can't play as much as when human was in their childhood. We have to develop the information system which is not perform information system itself but can help human to explore their natural instinct for playing, making fun and happiness when they interact with the information system. Virtual information system is the way to present playing and having fun atmosphere on working area.

## 1. Introduction

Information system an arrangement of people, data, processes, and information technology that interact to collect, process, store, and provide as output the information needed to support an organization [4]. Information system in organizations captures and manages data to produce useful information that supports an organization and its employees, customers, suppliers, and partners. The information system will need information technology support as they have made possible information system that could not exist without them. Moreover also depends on finding appropriate fit between the overall business goals, the information systems that help to fulfill those goals and the information technology on which the information system runs [1].

Information system in its development is influenced by some business drivers like [4] :
1) Globalization of the economy.
2) Electronic commerce and business.
3) Security and privacy.
4) Collaboration and partnership.
5) Knowledge asset Management
6) Continuous improvement
7) Total Quality Management
8) Business process redesign

Many organizations use information system as one of their ability to compete or gain competitive advantage. Organization increase their information system ability in order to compete with their competitor, watch out their new entrant, and substitutes and also give attention to bargaining with their supplier and customer as Michael Porter explained with his 5 forces competition [5][9]. In order to improve the information system, we need to make the system where employees, customers, suppliers and partners enjoy and easy to use when they use or get involve with our information system.

More than that we need something information system that can make them become more happy and fun when their make connection with our business. We need something that can make our business entities become loyal and pride when they have business connection with our information system. Happiness and fun are for all entities when they interact with the information system both internal and external entity. We need to combine between business and game in order to make human life become fun and happy when they do their business activity as the purpose when the computer technology was invented to make human life more be helped and easy to use to doing their daily activity with computer technology support.

Just remember game has to be fun [3]. The intermarriage between information system and game make human will be more interested to do their business activity as unreal as like playing game and as real as like to do their business activity.

## 2. Problem Formulation

Business entity like customer, supplier and employee are the entities which often have connection with the information system, especially for employee they have responsibilities to make sure the information system will run well. Employee from high level management to low level management have responsibility to deliver and maintain the best information system environment as an achievement to maintain their relation with customer and supplier and as value added to compete with the competitor [5]. High level management have a big challenge to enable and secure business strategy, which the business strategy drives the information system strategy, which in turn drives the information technology strategy[1][9].

The influences of business drivers and give the contribution for high level management for delivery best information system. Business driver Globalization of the economy will extend the competition be internationalized and push to support multiple languages, currency exchange rates, international trade regulation, different business cultures and practices. The growth of internet will expand the business process to implement e-commerce and e-business as other business driver. Business-to-Customer (B2C) and Business-to-Business (B2B) are the implementation of e-commerce and e-business.

Moreover business driver collaboration and partnership will drive cross-functional from many organization unit and outside organization even the competitors. Employees always come and go, the information system need knowledge and expertise for competitive advantage. So business driver knowledge management will drive to create and preserve knowledge.

Ultimately based on all that explanation, in this paper we try to find the answers from some questions like how to implement virtual information system as combining between information system and game in order to make happy information system? What have people known about virtual information system and what the differences with other technologies like virtual office?

# 3. Application architecture

Making easy how to implement virtual information system figure 1 picture out about what kinds of applications and database can be implemented on working area. Application architecture will be divided in three level application, they are:
1) Virtual application
2) OLTP (Online Transactional Processing)
3) OLAP (Online Analytical Processing)

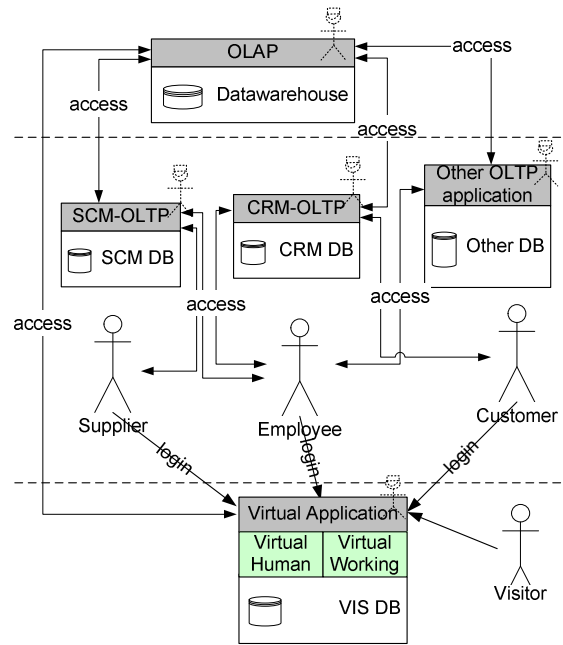

Fig 1. Application architecture

### 3.1 Virtual Application

The most important for application architecture is virtual application where this virtual application as the heart of virtual information system itself. All every actor's activities or conversation between actors or any knowledge will be saved on VIS DB database. Virtual application will be run as web application where people can access anywhere, anytime. Visitor who a real people that make first contact with the company. They no need register as an avatar. A visitor only visits some limit area which will be controlled by unreal avatar.

Virtual application will be divided as :
1) Virtual Human
   As a part of virtual application to manage an avatar both real and unreal personal.

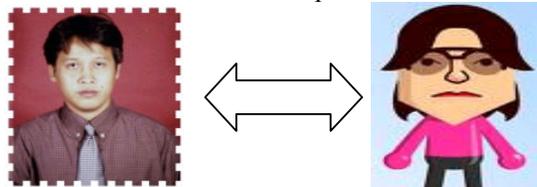

Real Human          Virtual Human
Fig 2. Real Human and Virtual Human

2) Virtual Working
   This Virtual Information system will be limit by physical working area. The virtual Information system just only show up what the reality in real working area and become the prohibition to show in virtual Information system something that never exist in real working area. On the other hand we can't build a new building or change the room position without to make changing in virtual Information system and every changing in reality working area must be recorded in virtual Information system. Hope everything look physically on working area will be shown up in virtual Information system and also everything look virtually in virtual Information system is a mapping from reality working area.

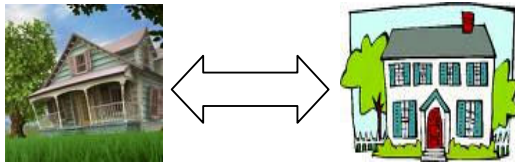

Real working area    Virtual Information system
Fig 3. Real working area and VIS

First of all entities like customer, supplier, employee from high level to low level management in Virtual Information system must making their own character as an avatar on Virtual application. When they want to access the virtual information system then they must login. All external entity like customer, supplier, and business partner no need to present itself physically in real working area, but their presence will be known by other people in virtual information system.

The way to login on the virtual information system could be many ways start from simple way by entering name and password on screen, or by sensor which the system can identify every each entity from their physical attendance. We can choose the technology to capture data entity with finger scan or eye scan to capture their personal data.

Specially for employee we can build the sensor which can detect automatically every employee when they come to their working area physically. Obviously there will be opportunity for employee to work remotely from other places outside working area and of course this kind of present work will be do it by enter their name and password on internet. Employee no need to present themselves physically, they can no attend to their office to doing their jobs, serve their customer and presence virtually as an avatar personal.

There is also possibility to hire blind, dumb, deaf and other physical defect as employee. In one hand will help employee with physical defect work remotely and the other hand will give an opportunity for company to hire employee more lower cost both lower in pay per hour and provide places for working.

What virtual application can do:

1) Input/edit avatar for represent each of entity character
2) There are 2 kinds of avatar
   a. Real avatar
      An avatar which represent as all entities like customer, supplier and real employee from high level management to low level management. Every entity will only have one avatar and every avatar can represent one man or as unreal avatar. In figure 1 real avatar is represented as picture:

      Could be person will meet with other people who have well known before and can talk each others by virtually. Also there will happen to make a friendship with someone who always meet when online on Virtual information system and possibility to become more closely and have a closed relationship. Every entity will know the other entities who online and there will be no limit between present as physically and present as virtual.
   b. Unreal avatar
      An avatar which represent as visual people where act as visual employee. This is become value added for company to hire employee without to pay them cause they are virtual. The employer doesn't need many employee, they can have virtual employee [7] where they don't have pay their salary, benefit and any employee's need. In figure 1 unreal avatar is represented as picture :

      In each core application there will be one unreal avatar which responsible for each core application and could be designed along with company needed to assign how many unreal avatar.

   Need some regulation is there any requirement to differentiate between real and unreal avatar.
3) Communication between avatar both real and unreal will be recorded and will sharpen intelligence each of avatar, specially for unreal avatar. Need some regulation if is there any requirement to limit any personal activity from recording process.

   The owner will have information about their daily employee's activity. If Every movement from each person will be recorded in the Virtual Information system and need some limitation not to record every people's personal activity like bathroom activity. Physically people will more be stress when their daily activity will be recorded in virtual information system include their personal activity like bathroom activity.

Sensor will be needed to know where the people position physically and will transform to Virtual Information system where they are virtually. Here we need like smart card for cell phone which has been stored with people personal identification. But we need some extend research to know what will happen if they lose their smart card.

4) If someone physically die then by virtually they will be exist and by agreement between employee, employee's family and company then the dead people will be raised as an unreal avatar character in virtual information system. No surprise if we can still speak with their avatar even they have already died.
5) By saving all then much knowledge will be saved and will help the next people by all the knowledge which has been saved. By this knowledge we can extend intelligence information system as part of OLAP which can predict the future by all that knowledge. All the data in VIS DB database will be extracted to datawarehouse [10].
6) Virtual application can make contact with external virtual application/ external virtual human like Ramona at www.kurzweilAI.net or Alice at www.alicebot.org. Also can make connection between one virtual information system with the other virtual information systems. We need some standardization to make all virtual world will work together. Some regulation will be needed and controlled by approved association.
7) For the next virtual information system will become more intelligent, can learn and do some task without supervision and become intelligent agent.[11] There will be no dream if one avatar will talk to another avatar without human touch. Our avatar will learn by themselves and even more clever than us [7].
8) Speech recognition and voice generation will be extended on virtual human in order to accurate language translation between many languages both verbal and textual [8][11]. People from many cultures and languages will interact with their own language.

### 3.2 OLTP (Online Transactional Processing)

OLTP or can be called TPS (Transactional Processing System) as applications for daily transactions which are run by low level management. Level OLTP is designed become 3 core application are : SCM-OLTP, CRM-OLTP and other OLTP applications. As Porter explained with his 5 forces competition to give attention for supplier and customer's bargaining then we need to build applications which can serve them as external entities [5][9]. For serving supplier we build SCM (Supply Chain Management) and for serving customer we build CRM (Customer Relationship Management).

For performance each of databases will be set in each of core application. Design for this level will be along with business needed and infrastructure architecture. Before access these OLTP applications, each of entity must be login first in virtual application. All the entity's activities when online on virtual information system will be recorded. From each core application the entity can access the OLAP reports. By scheduling data from each OLTP database will be extracted to datawarehouse [10][11].

For the implementation better if each of core application will be designed along with physical room design. Physically each core application will have its room and reverse. When Supplier login on virtual information system they will show up the avatar on SCM room and also with the customer when they login they will show up avatar on CRM room. For employee they will show up avatar at the place where they are assigned.

### 3.3 OLAP (Online Analytical Processing)

Applications which usually are run by middle and high level management but for improving will be better if these applications will also be delivered for low level management and external entities like customer or supplier in limited environment [10]. OLAP whatever the names are the same and could be DSS (Decision Support System), Business Intelligence, Data Mining, Expert System, EIS (Executive Information system), knowledge Management System.

So no matter what the names then depend on the organization what the technology that will be used but one thing that make a sure OLAP will deliver the best performance information deliverance. Easy to access, fast time to deliver, can extend to data sources and many alternatives reports.

Figure 1 above as one of the suggestive implementation and the other developments can be arranged by each business institution as the organization wanted. Application and database design can be designed along with hardware design and also with other software requirements like operating system or security system. Finally performance will be the most important thing to deliver best virtual information system implementation design.

## 4. The Differentiation

After design the application infrastructure for virtual information system then we need to know what people have known about virtual information system. After searching on internet then we have some information such as:

1) In 1997 Song L and Nagi R wrote about virtual information system but what they had researched about the requirement, design and implementation issues of agile manufacturing information system by present AMIS (Agile Manufacturing Information System framework). AMIS as an open system architecture was used to merge some manufacturing information which have each expertise, partners from different companies, who collaborate with each other to design and manufacture high quality and customized products.[12][13].
2) In 2000 Padma, V.U. and Harish B wrote about case study on a virtual information system for a group of institutions in Manipal, India. The VIS was defined as "A network of different library systems that come together to exploit or transit the information available in their premise or they can also be defined as Information System Partners.[14] One of their future strategy for virtual information system is to develop multimedia sources.
3) Virtual Information System Store at http://stores.ebay.com/Virtual-Information-Systems . Ebay provide the Internet platforms of choice for global commerce, payments and communications. With Virtual Information System Store, Ebay helps their customer to selling by online business and helps their buyer according to specific needs and interest. People around the world differences in language, culture and economic status explore, learn, shop, share and talk with each other.
4) Infomaster project which has developed at Computer Science Department, Standford University. Infomaster is a virtual information system that allows users to access a variety of heterogeneous, distributed information from multiple perspectives. Infomaster accesses information stored in databases or knowledge bases using Agent Communication Language (ACL). Infomaster is an operational system that currently presents information on rental housing in the San Francisco Bay Area and on people at Stanford [6].
5) In 2006 Amberg, M. & Fischer S wrote about web-based Aptitude test for student Guidance where they proposed virtual advice as information system for freshmen guidance with standardized web-based aptitude tests by adding virtual environment like 360 degree images, movie sequences and photo.[16].
6) Fairer-Wessels with his idea about the development of leisuredigest.com, a South African tourism website wrote that A virtual information system is a specific type of information system that is based on the WWW. According to his written it is virtual in the sense that it is not book or paper-based although information on the Web can be downloaded onto paper or hard format. The advantage of virtual information is that it can be updated on a daily basis without the cost involved in updating traditional paper-based publications, as well as the advantage of timeliness, especially in the case of tourist information that is often obsolete before it is printed.[17]
7) In 1995 Lau, J. wrote about virtual libraries for less developed country where virtual information system as 3D information system or virtual reality information system to come up with alive pictures, sound and smell of real life. [15]. He wrote that virtual information system offer libraries the possibility of providing tools with full multimedia interacting power, besides the traditional printed and electronic services.

Padma with his concept for Manipal institutions and Song with his concept about virtual information system for agile manufacturing are basic foundation to deliver the information system as virtual with networking without web support. Web and multimedia is just as future suggestion for their virtual information system.

Others like ebay, infomaster project, web-based Aptitude Test, Fairer-Wessels' leisuredigest.com and Lau's virtual libraries start to use web application for their virtual information system. They proposed using web technology with multimedia support such as sounds, pictures, animation and movies.

Virtual application with virtual working and virtual human in my writing as a new approach for the new era for the meaning of virtual information system, where the information system will be declared by virtually as the mankind do their natural instinct to play, live happy in this world. Mankind will do their business activity as real as they are live, as real as when they play and become clever and their knowledge will be live after when they were dead. There will be an virtual human both real avatar or unreal avatar who will manage the virtual information system and there will be an virtual working area which can be approved by mankind for their own good to make easy life in this world, happy and full of joy.

## 5. Conclusion

There will be many abuse of right and manipulation on the implementation but it always happen in first runtime, more than that if we apply on internet but next will make the system more secure. For performance the virtual information system implementation will need high performance infrastructure.

Virtual application with virtual human and virtual working become the most interesting, will need some future research from present research and specially for the implementation what is the right software application which apply them and what is the right technology to make them will be implemented.